\documentclass[10pt,letterpaper,twocolumn]{article}
\usepackage[latin9]{inputenc}
\usepackage{mathrsfs}
\usepackage{amsmath}
\usepackage{amssymb}
\usepackage{graphicx}
\usepackage[unicode=true,pdfusetitle,
 bookmarks=false,
 breaklinks=true,pdfborder={0 0 1},backref=section,colorlinks=false]
 {hyperref}

\makeatletter

\providecommand{\tabularnewline}{\\}

\usepackage{wacv}\usepackage{times}\usepackage{epsfig}

\wacvfinalcopy 

\def\wacvPaperID{295} 

\ifwacvfinal\fi

\makeatother

\begin{document}

\title{Learning to Recognize Objects by Retaining other Factors of Variation} 

\author{Jiaping Zhao, Chin-kai Chang,
 Laurent Itti \\
 University of Southern California\\
\texttt{\small{}{}\{jiapingz, chinkaic, itti\}@usc.edu}{\small{}{} }}
\maketitle
\begin{abstract}
Most ConvNets formulate object recognition from natural images as a single task classification problem, and attempt to
learn features useful for object categories, but invariant to other factors of variation such as pose and
illumination. They do not explicitly learn these other factors; instead, they usually discard them by pooling and
normalization.  Here, we take the opposite approach: we train ConvNets for object recognition by retaining other factors
(pose in our case) and learning them jointly with object category. We design a new multi-task leaning (MTL) ConvNet,
named disentangling CNN (disCNN), which explicitly enforces the disentangled representations of object identity and
pose, and is trained to predict object categories and pose transformations. disCNN achieves significantly better object
recognition accuracies than the baseline CNN trained solely to predict object categories on the iLab-20M dataset, a
large-scale turntable dataset with detailed pose and lighting information. We further show that the pretrained features
on iLab-20M generalize to both Washington RGB-D and ImageNet datasets, and the pretrained disCNN features are
significantly better than the pretrained baseline CNN features for fine-tuning on ImageNet.
\end{abstract}

\section{Introduction}


Images are generated under factors of variation, including pose, illumination etc. Recently, deep ConvNet
architectures learn rich and high-performance features by leveraging millions of labelled images. They have achieved
state-of-the-art object recognition performance. Contemporary CNNs, such as AlexNet \cite{krizhevsky2012imagenet}, VGG
\cite{simonyan2014very}, GoogLeNet \cite{szegedy2015going} and ResNet \cite{he2015deep}, pose object recognition as a
single task learning problem, and learn features that are sensitive to object categories but invariant to other nuisance
information (e.g., pose and illumination) \cite{soatto2016visual} as much as possible. To achieve this, current CNNs usually stack several
stages of subsampling/pooling \cite{lecun1998gradient} and apply normalization operations
\cite{krizhevsky2012imagenet,ioffe2015batch} to make representations invariant to small pose variations and illumination
changes. However, as argued by Hinton et al \cite{hinton2011transforming}, to recognize objects, neural networks should
use ``capsules'' to encode both identity and other instantiation parameters (including pose, lighting and shape
deformations). In \cite{bengio2009learning,reed2014learning}, authors argue as well that image understanding is to tease
apart these factors, instead of emphasizing one and disregarding the others.

In this work, we formulate object recognition as a multi-task learning (MTL) problem by taking images as inputs and
learning both object categories and other image generating factors (pose in our case) simultaneously.  Thanks to the
availability of both identity and 3D pose labels in the iLab-20M dataset of 22 million images of objects shot on a
turntable, we use object identity and pose during training, and then investigate further generalization to other
datasets which lack pose labels (Washington RGB-D and ImageNet). Contrary to the usual way to learn representations
invariant to pose changes, we take the opposite approach by retaining the pose information and learning it jointly with
object identities during the training process.

\begin{figure}[!htbp]
\begin{centering}
\includegraphics[width=0.48\textwidth]{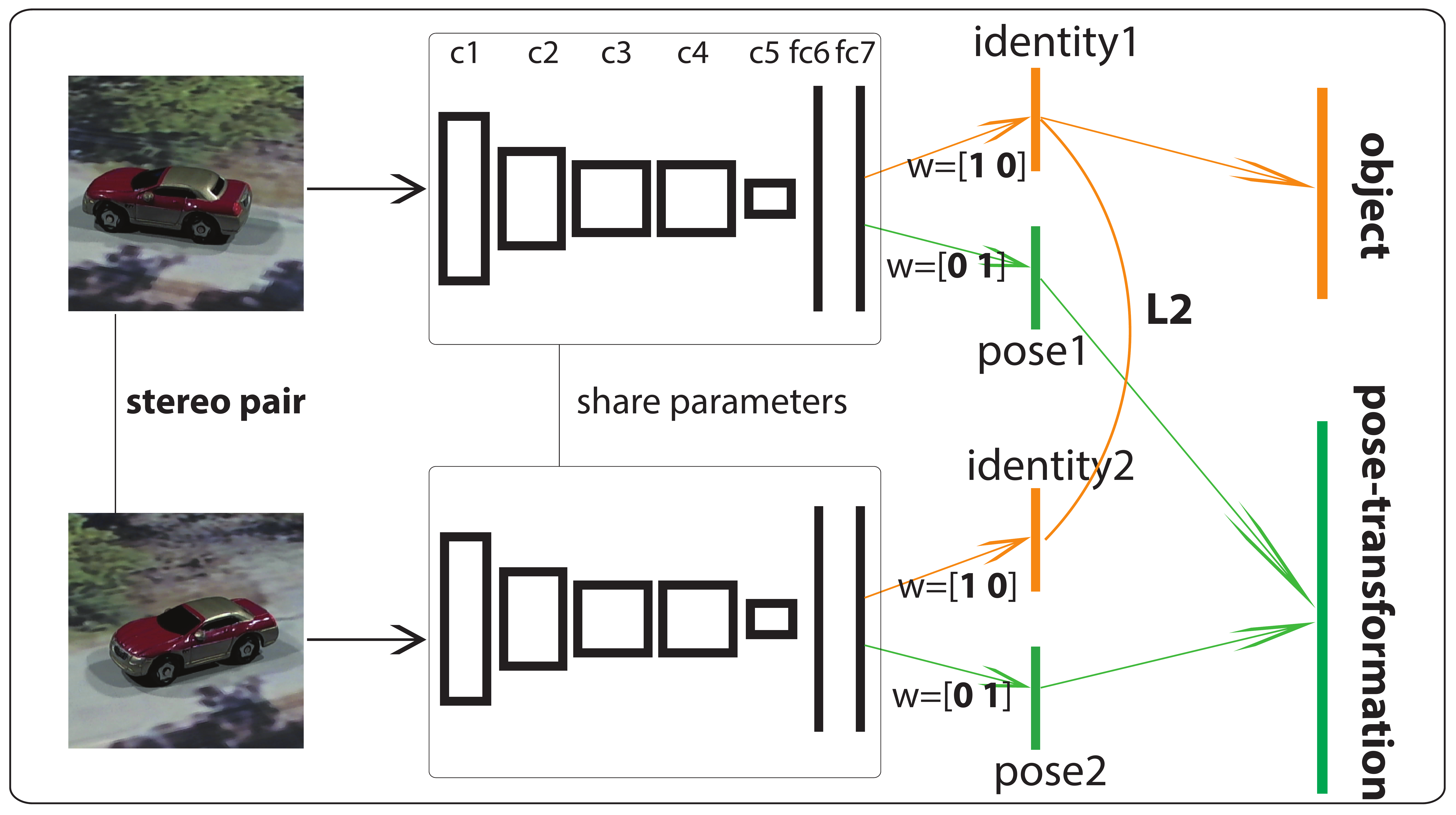}
\par\end{centering}

\centering{}\caption{\label{fig:Architecture}Architecture of our disentangling CNN (disCNN).
disCNN is a two-streams CNN, which takes in an image pair
and learns to predict the object category and the pose transformation
jointly. In experiments, we use AlexNet in both streams to extract
features, and explicitly partition the top layer representations fc7
into two groups: identity and pose. We further enforce two identity
representations to be similar, and one identity representation
is used for object category prediction,
and two pose representations are concatenated to predict the pose
transformation.}
\end{figure}

We leverage the power of ConvNets for high performance representation learning, and build our MTL framework on
it. Concretely, our architecture is a two-streams ConvNet which takes a pair of images as inputs and predicts both the
object category and the pose transformation between the two images. Both streams share the same CNN architecture (e.g.,
AlexNet) with the same weights and the same operations on each layer.  Each stream independently extracts features from
one image. In the top layer, we explicitly partition the representation units into two groups, with one group
representing object identity and the other its pose. Object identity representations are passed down to predict object
categories, while two pose representations are concatenated to predict the pose transformation between images
(Fig. \ref{fig:Architecture}). By explicitly partitioning the top CNN layer units into groups, we learn the ConvNet in a
way such that each group extracts features useful for its own task and explains one factor of variation in the image. We
refer our architecture as disentangling CNN (disCNN), with disentangled representations for identity and pose.

During training, disCNN takes a pair of images as inputs, and
learns features by using both object categories and pose-transformations
as supervision. The goal of disCNN is to recognize objects,
therefore, in test, we take only one stream of the trained disCNN,
use it to compute features for the test image, and only the identity
representations in top layer are used and fed into the object category
layer for categorization. In other words, pose representations are
not used in test, and the pose-transformation prediction task in the
training is auxiliary to the object recognition task, but essential
for better feature learning.




\section{Related work}



ConvNets: over the past several years, convolutional neural networks \cite{lecun1998gradient} have pushed forward
the state-of-the-art in many vision tasks, including image classification
\cite{krizhevsky2012imagenet,simonyan2014very,szegedy2015going,he2015deep}, object detection
\cite{sermanet2013overfeat,girshick2014rich}, image segmentation \cite{chen2014semantic,long2015fully}, activity
recognition \cite{simonyan2014two,gkioxari2015contextual}, etc. These tasks leverage the power of CNNs to learn rich
features useful for the target tasks, and \cite{agrawal2015learning} show features learned by CNNs on one task can be
generalized to other tasks. We aim to learn feature representations for different image generating factors, and we
employ ConvNets as our building base.

Multitask learning: several efforts have explored multi-task learning
using deep neural networks, for face detection, phoneme recognition,
and scene classification \cite{seltzer2013multi,zhang2014facial,zhang2014improving,huang2013multi}.
All of them use a similar linear feed-forward architecture, with all
task label layers directly appended onto the top layer.
In the end, all tasks in these applications share the same representations.
More recently, Su et al \cite{su2015render} use a CNN to estimate the
camera viewpoint of the input image. They pose their problem as MTL
by assuming that viewpoint estimate is object-class-dependent, and
stack class-specific viewpoint layers onto the top of CNN. Zhao and Itti \cite{zhao2016improved} formulated object recognition in a CNN-based MTL framework, and the top layer representations are shared among tasks as well. Our work
differs from the above in that: we use two-stream CNNs and we explicitly
partition the top layer representation into groups, with each group
representing one task; therefore we have task-exclusive representations
while in above works, all tasks share the same top layer representations.

Disentangling: As argued by Bengio \cite{bengio2009learning}, one of the key challenge to understanding images is to
disentangle different factors, e.g. shape, texture, pose and illumination, that generate natural images.
Reed et al \cite{reed2014learning} proposed the disentangling Boltzmann Machine (disBMs), which augments the regular RBM
by partitioning the hidden units into distinct factors of variation and modelling their high-order interactions. In \cite{zhu2014multi},
the authors build a stochastic multi-view perceptron to factorize the face identity and its view
representations by different sets of neurons, in order to achieve view-invariant face recognition. Our work is
similar to the above two in that we explicitly partition the representations into distinct groups to force different
factors disentangled; however, our model is deterministic and scales to large datasets, while the above methods are
restricted to small datasets and often require expensive sampling inferences.

Dosovitskiy et al \cite{dosovitskiy2015learning} proposed to use CNN to generate images of objects given object style,
viewpoint and color. Their model essentially learns to simulate the graphics rendering process, but does not directly
apply to image interpretation. Kulkarni et al \cite{kulkarni2015deep} presented the Inverse Graphics Network (IGN), an
encoder-decoder that learns to generate new images of an object under varying poses and lighting. The encoder of IGN
learns a disentangled representation of transformations including pose, light and shape. Yang et al
\cite{yang2015weakly} proposed a recurrent convolutional encoder-decoder network to render 3D views from a single
image. They explicitly split the top layer representations of the encoder into identity and pose units. Our work is
similar to \cite{kulkarni2015deep,yang2015weakly} by using distinct units to represent different factors, but we differ
in that: (1) our architecture is a MTL CNN which maps images to discrete labels, while theirs are autoencoders mapping
images to images; (2) our model directly applies to large numbers of categories with complex images, but
\cite{kulkarni2015deep,yang2015weakly} only tested their models on face and chair datasets with pure backgrounds.

Our work is most similar to \cite{agrawal2015learning}, which shows that freely available egomotion data of mobile
agents provides as good supervision as the expensive class-labels for CNNs to learn useful features for different vision
tasks. Here, we use stereo-pairs of images as inputs to learn the camera motions as well, however, we are different in:
(1) our architecture is a MTL framework, in which the task of camera-motion (pose-transformation) prediction serves as
an auxiliary task to help object recognition; (2) our network is more flexible, which could take in both one image or a
stereo-pair; (3) our MTL disCNN learns much better features for object-recognition than the baseline CNN using only
class-label as supervision, while their single task two-streams CNNs only learn comparable features.


%

\section{Method\label{sec:Method}}

Object identity is here defined to be the identity of one instance. Distinct instances, no matter whether they belong to the same class, have different object identities. Object pose refers to the extrinsic parameters of the camera taking the image, but given a single natural image taken by a consumer camera, it is hard to obtain the camera extrinsics, therefore the ground-truth object poses are expensive and sometimes impossible to collect in real cases. Camera extrinsics are known when we render 2D images from 3D models\cite{su2015render}, but rendered 2D images are very different from natural images. Although single camera extrinsics are hard to get in real cases, the relative translation and orientation (a.k.a camera motion), represented by an essential matrix $E$, between a pair of images is relatively easier to compute, e.g., for calibrated cameras, first find 8 pairs of matched points, then use the normalized 8-point algorithm \cite{hartley2003multiple} to estimate. The camera motion between an image pair captures the pose transformation between objects in the two images. We use the relative pose transformation between objects instead of absolute object pose, as supervision.  In the following, we use ``pose transformation'' and ``camera motion'' interchangeably.

Our system is designed to estimate any numeric pose transformations, but in experiments, we have a limited number
of camera-pairs, with motion between each pair fixed. Therefore, we could further discretize the pose transformation
using the fact that every image-pair taken under the same camera-pair has the same pose transformation, and the number
of the camera-pairs determines the number of discrete pose-transformations.  In this way, ``pose transformation''
estimation is transformed into a classification problem - classifying which camera-pair took the image-pair, with the
number of labels equal to the number of camera-pairs.



\subsection{Network Architecture}

Our ultimate goal is to learn object identity representations for
object recognition, but we further simultaneously learn the object
pose transformation as an auxiliary task. Building a ConvNet that
can predict the pose transformation between a stereo-pair of images
is straightforward: the ConvNet should take the pair as input, after
several layers of convolutions, it produces an output which assigns
a probability to each camera-pair under which that image-pair could
be taken. But note that the image-pair contains the same object instance
taken under different camera viewpoints, we wish to learn an object
identity representation, such that the same pair should have as similar
object identity representations as possible.

We build a two-stream CNN architecture shown in Fig.\ref{fig:Architecture}, named disentangling CNN (disCNN). Each
stream is a ConvNet independently extracting features from one image, and both ConvNets have the same architecture and
share the same weights. Here we use AlexNet \cite{krizhevsky2012imagenet} as the ConvNet, but with faster GPUs one could
use VGG \cite{simonyan2014very} and GoogLeNet \cite{szegedy2015going} as well. After getting fc7 representations, we
explicitly partition the fc7 units into two groups, with one group representing object identity and the other
representing object pose in a single image. Since object instances in an image pair are the same, we enforce the two
identity representations to be similar by penalizing their $\ell_{_{2}}$-norm differences, i.e. $\parallel
id_{1}-id_{2}\parallel_{2}$, where $id_{1}$ and $id_{2}$ are identity representations of two stereo images. One identity
representation (either $id_{1}$ or $id_{2}$) is further fed into object-category label layer for object-category
prediction. Two pose representations, $pose_{1}$and $pose_{2}$, are fused to predict the relative pose transformation,
i.e., under which camera-pair the image-pair is taken. Our objective function is therefore the summation of two soft-max
losses and one $\ell_{2}$ loss:

\vspace{-1em}

\begin{equation}
\begin{split}
\mathcal{L}= & \quad\mathcal{L}(object)+ \\
 & \lambda_{1}\mathcal{L}(pose\; transformation)+ \\
 & \lambda_{2}\parallel id_{1}-id_{2}\parallel_{2}\label{eq:loss-objective}
\end{split}
\end{equation}


We follow AlexNet closely, which takes a $227\times227$ image as input, and has 5 convolutional layers and 2 fully
connected layers. ReLU non-linearities are used after every convolutional/fully-connected layer, and dropout is used in
both fully connected layers, with dropout rate 0.5. The only change we make is to change the number of units on both fc6
and fc7 from 4096 to 1024, and one half of the units (512) are used to represent identity and the other half to
represent pose.  If we use abbreviations Cn, Fn, P, D, LRN, ReLU to represent a convolutional layer with n filters, a
fully connected layer with n filters, a pooling layer, a dropout layer, a local response normalization layer and a ReLU
layer, then the AlexNet-type architecture used in our experiments is:
C96-P-LRN-C256-P-LRN-C384-C384-C256-P-F1024-D-F1024-D (we omit ReLU to avoid cluttering). If not explicitly mentioned,
this is the baseline architecture for all experiments.

\begin{figure*}[!htbp]
\begin{centering}
\includegraphics[width=0.8\textwidth]{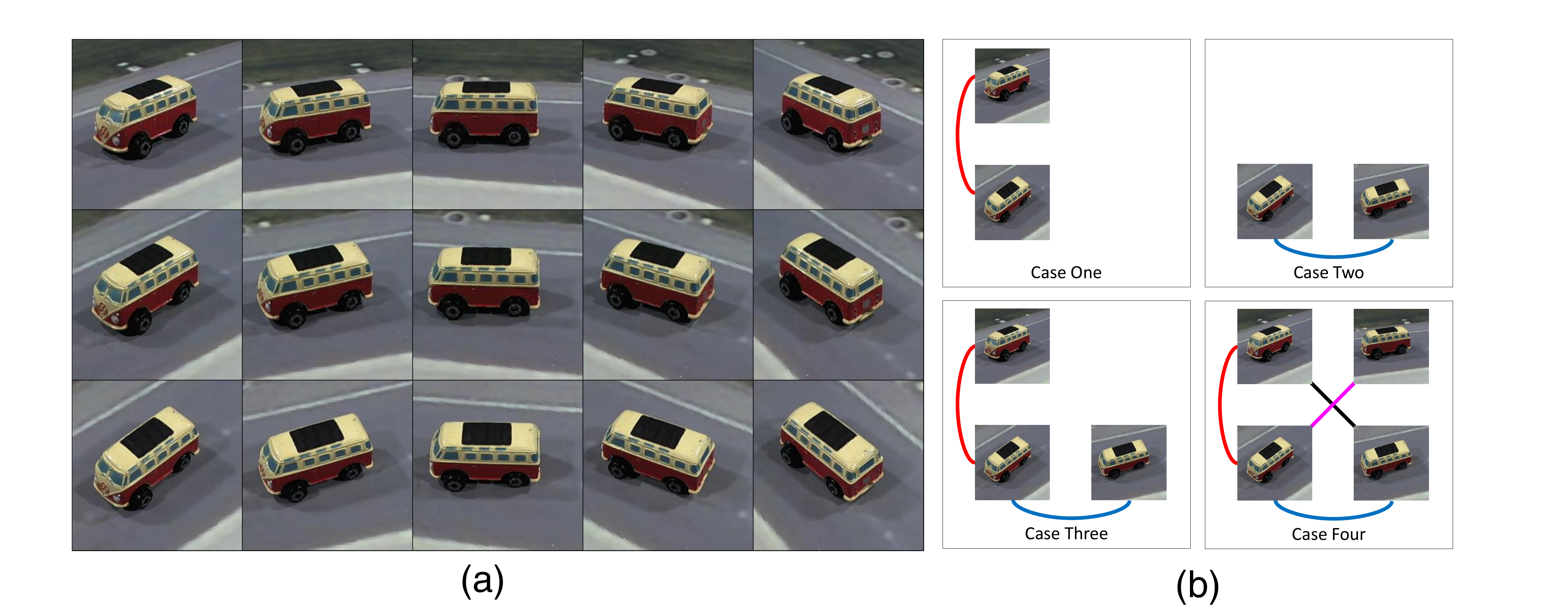} 
\par\end{centering}

\centering{}\caption{\label{fig:Exemplar-iLab-20M-images-camera-pairs}Exemplar iLab-20M
images and camera pairs. (a) images of the same object instance taken
by different cameras under different rotations, each row is taken
under the same camera under different rotations, and each column is
taken by different cameras under the same rotation; (b) camera pairs
used in experiments.}
\end{figure*}

Notes: (1) the proposed two-stream CNN architecture is quite flexible in that: it could either take a single image or an
image-pair as inputs. For a single image input, no pose transformation label is necessary, while for an image-pair
input, it is not required to have an object-category label. For a pair of images without the object label, its loss
reduces to two terms: $\lambda_{1}\mathcal{L}(pose-transformation)+\lambda_{2}\parallel id_{1}-id_{2}\parallel_{2}$, the
soft-max loss of the predicted pose-transformation and the $\ell_{2}$ loss of two identity representations. Given a
single image with a object label, the loss incurred by it reduces to only one term: the soft-max loss of the predicted
category label $\mathcal{L}(object)$.

(2) Scaling the same image-pair by different scales does not change its pose transformation label. In our case, each
camera-pair has a unique essential matrix (up to some scale), and defines one pose-transformation label. By up/down
scaling both images in a pair, the estimated essential matrix differs only by a scale factor. Since the essential matrix
estimated from the raw image-pair is already uncertain up to a scale factor (e.g. using the eight-point method for
estimation \cite{hartley2003multiple}), the essential matrix estimated from the scaled pairs is equivalent to that
estimated from the raw pair.  This is useful when objects have large scale differences: we could scale them differently
to make them have similar scales (see experiments on Washington RGB-D dataset).

\section{Experiments\label{sec:Experiments}}

In experiments, we first show the effectiveness of disCNN for object
recognition against AlexNet on both iLab-20M and Washington RGB-D datasets.
We further demonstrate that the pretrained disCNN on the iLab-20M dataset
learns useful features for object recognition on the ImageNet dataset
\cite{deng2009imagenet}: a AlexNet initialized with disCNN weights
performs significantly better than a AlexNet initialized with random
Gaussian weights.

\subsection{iLab-20M dataset}

The iLab-20M dataset \cite{Borji_2016_CVPR} is a controlled, parametric dataset collected by shooting
images of toy vehicles placed on a turntable using 11 cameras at
different viewingpoints. There are totally 15 object categories with
each object having 25$\sim$160 instances. Each object instance was
shot on more than 14 backgrounds (printed satellite images), in a
relevant context (e.g., cars on roads, trains on railtracks, boats
on water). In total, 1,320 images were captured for each instance
and background combinations: 11 azimuth angles (from the 11 cameras),
8 turntable rotation angles, 5 lighting conditions, and 3 focus values
(-3, 0, and +3 from the default focus value of each camera). The complete
dataset consists of 704 object instances, with 1,320 images per object-instance/background
combination, or almost 22M images.

Training and test instances: we use 10 (out of 15) object categories
in our experiments (car, f1car, helicopter, plane, pickup, military car, monster, semi, tank and van),
and, within each category, we randomly choose 3/4 instances as training
and the remaining 1/4 instances for testing. Under this partition,
instances in test are never seen in training.

Image-pairs: we only take images shot under one fixed lighting condition (with all 4 lights on) and camera focus (focus
= 0), but all 11 camera azimuths and all 8 turntable rotations as training and test images, equivalent to 88 virtual
cameras on a semi-sphere.  In principle, we can take image-pairs taken under any camera-pairs (e.g. any pair from
$C_{88}^{2}$ combinations), however, one critical problem is that image-pairs taken under camera-pairs with large
viewpoint differences have little overlap, which makes it difficult, or even impossible to predict the
pose-transformation (e.g., difficult to estimate the essential matrix). Therefore, in experiments, we only consider
image-pairs taken by neighboring-camera pairs. All image-pairs shot under a fixed camera-pair share the same
pose-transformation label, and finally the total number of pose-transformation labels is equal to the number of
camera-pairs. In experiments, we consider different numbers of camera-pairs, and evaluate the influence on the
performance of disCNN.

\begin{table}[!htbp]
\centering{}%
\begin{tabular}{|c|c|c|c|c|}
\hline 
\# of camera pairs  & 7  & 11  & 18  & 56 \tabularnewline
\hline 
\hline 
AlexNet  & 79.07  & 78.89  & 79.60  & 79.25 \tabularnewline
\hline 
disCNN  & \textbf{81.30}  & \textbf{83.66}  & \textbf{83.60}  & \textbf{83.66} \tabularnewline
\hline 
\end{tabular}\caption{\label{tab:Object-recognition-iLab-20M}Object recognition accuracies
($\%$) of AlexNet and disCNN on the iLab-20M dataset. disCNN consistently
outperforms AlexNet under different numbers of camera pairs, showing the advantage of jointly learning object identity
and its pose. We see as well: disCNN performs better when more camera-pairs
are used, e.g., the performance of disCNN increases by $2\%$ when
$\geq$11 camera pairs are used, compared with 7 camera pairs. }
\end{table}

Fig. \ref{fig:Exemplar-iLab-20M-images-camera-pairs} shows images of one instance shot under different cameras and
rotations: each row is shot by the same camera under different turntable rotations, and each column is shot by different
cameras under the same turntable rotation. In experiments, we use different numbers of camera-pairs as supervision,
therefore, only take image-pairs shot under the chosen camera-pairs as training. Case one
(Fig. \ref{fig:Exemplar-iLab-20M-images-camera-pairs} (a) topleft): we take two neighboring cameras as one camera-pair
(we skip 1 camera, i.e., $C_{i}-C_{i+2}$ is a camera-pair), resulting in 7 camera-pairs, therefore 7 pose-transformation
labels. Image pairs taken by the same camera-pair under different rotations share the same pose-transformation
label. Case two (Fig. \ref{fig:Exemplar-iLab-20M-images-camera-pairs} (b) topright): two images taken by one camera
under two adjacent rotations ($(C_{i}R_{j},\, C_{i}R_{j+1})$) can be imagined to be taken by a pair of virtual cameras,
resulting in 11 camera-pairs with 1 pair referring to one camera under two adjacent rotations. Case three (Fig.
\ref{fig:Exemplar-iLab-20M-images-camera-pairs} (c) bottomleft): we combine 7 camera-pairs in case one and 11
camera-pairs in case two, and a total of 18 camera pairs. Case four
(Fig. \ref{fig:Exemplar-iLab-20M-images-camera-pairs} (d) bottomright): in addition to take image-pairs taken under
neighboring cameras (the same rotation) and neighboring rotations (the same camera), we further take diagonal
image-pairs taken under neighboring-cameras and neighboring-rotations (i.e., $(C_{i}R_{j},\, C_{i+1}R_{j+1})$ and
$(C_{i}R_{j+1},\, C_{i+1}R_{j})$). At last we have 56 camera-pairs.  By taking image-pairs from the chosen camera-pairs,
we end up 0.42M, 0.57M, 0.99M and 3M training image-pairs in 4 cases respectively.  After training, we take the trained
AlexNet-type architecture out and use it to predict the object category of a test image. We have a total of 0.22M test
images by split.

Implementation details: Since we have prepared training pairs for disCNN, we use the left images of training pairs as
the training data for AlexNet. Therefore AlexNet and disCNN have the same number of training samples, with one image in
AlexNet corresponding to an image pair in disCNN (Note: duplicate training images exist in AlexNet).  To do a fair
comparison, we train both AlexNet and disCNN using SGD under the same learning rate, the same number of training epochs
and the same training order within each epoch. We set $\lambda_{1}=1$ and $\lambda_{2}=0.1$ in the objective function
\ref{eq:loss-objective} of disCNN. Practically, $\lambda_{1}$ and $\lambda_{2}$ are set such that the derivatives of
three separate loss terms to the parameters are at a similar scale. Both AlexNet and disCNN are trained for 20 epochs
under 4 cases. The initial (final) learning rate is set to be 0.01 (0.0001), which is reduced log linearly after each
epoch.  The ConvNets are trained on one Tesla K40 GPU using the toolkit \cite{vedaldi2015matconvnet}.

\noindent \textbf{Notes}: although disCNN is trained on the stereo pairs, and AlexNet is only trained on the left images, we have to emphasize that: for each stereo-pair, the left image is also the right image of another stereo-pair, and the right image is also the left image of some other stereo-pair (see Fig. \ref{fig:Exemplar-iLab-20M-images-camera-pairs} (b) for how we prepare the stereo pairs). This means, for the input stereo-pairs, all right images are actually a shuffled version of all left images. Therefore, the stereo-pairs do not contain new images, which do not belong to all left images. So both disCNN and AlexNet are actually trained on the same images, and results are comparable.

\begin{figure*}
\centering{}\includegraphics[width=1\textwidth]{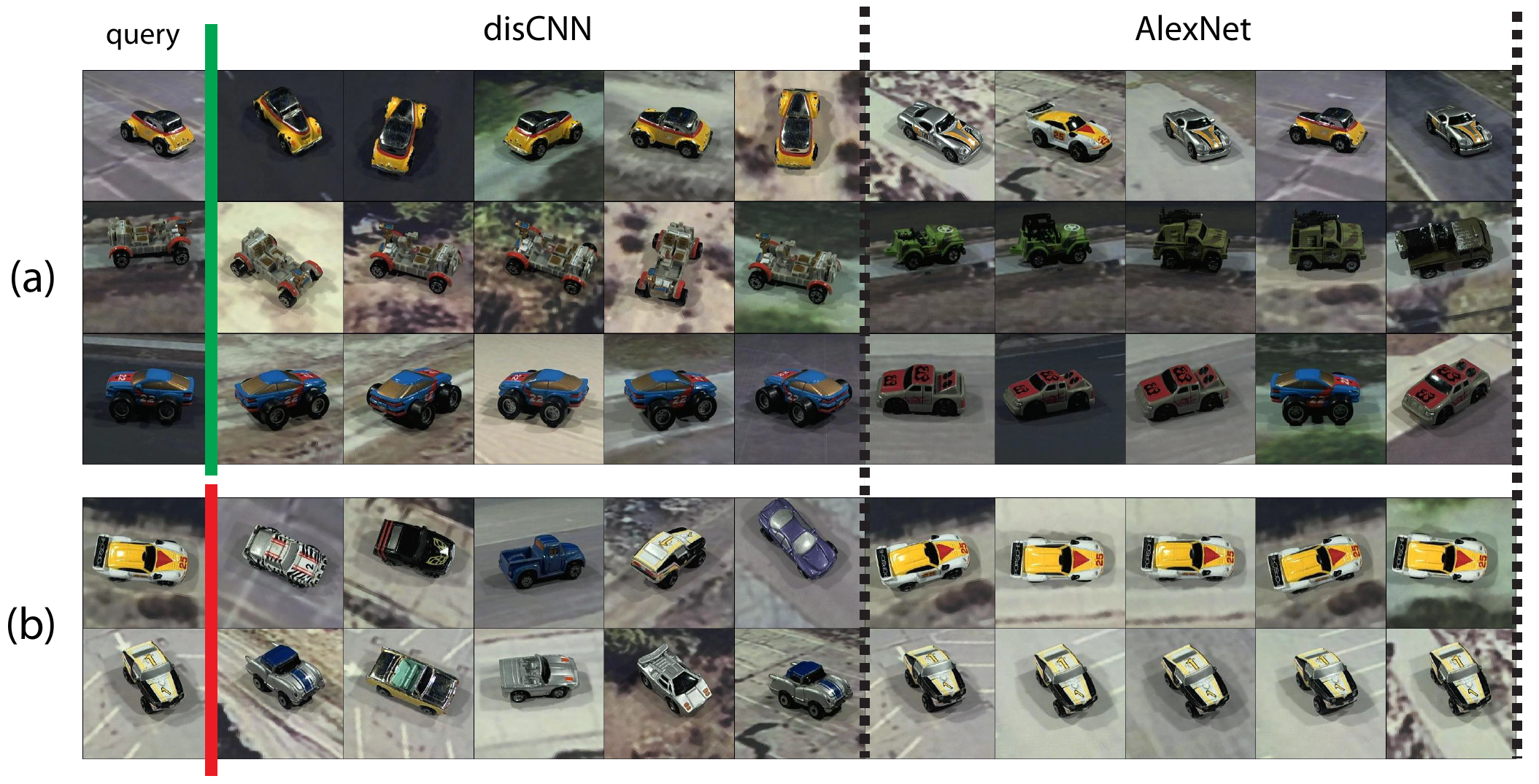}
\caption{\label{fig:disentangling}Examples of k nearest neighbors of query
images. Images are represented by fc7-identity (disCNN, 512D)
and fc7 (AlexNet, 1024D) features, and then 5 nearest neighbors are
searched based on $\ell_{2}$ distances in the representation spaces.
On each row, the 1st image is the query image, and the next 5 (the
last 5) images are retrieved nearest neighbors by disCNN and AlexNet.
In group (a), disCNN always returns the same instance but under different
poses as the nearest neighbors, but AlexNet fails to retrieve the
same instance, instead it returns instances with different identities
but similar poses. In group (b), although disCNN fails to retrieve
the right instance, it does find instances with similar shapes to
the query image. In this case, AlexNet retrieves the correct instances
with the same identity, but again, the poses of the retrieved images
are very similar to the query one. This result shows disCNN
disentangles identity from pose, to some extent.}
\end{figure*}

%

\textbf{Results}: the object recognition performances are shown in Table \ref{tab:Object-recognition-iLab-20M}.
We have the following observations: (1) disCNN consistently outperforms
AlexNet under different numbers of camera pairs, with the performance
gain up to $\sim4\%$; (2) when we have more camera-pairs, the performance
gap between disCNN and AlexNet widens, e.g., $\sim4\%$ gain under
11,18,56 camera pairs compared with $\sim2\%$ gain under 7 camera
pairs. One potential reason is that when more camera pairs are used,
more views of the same instance are available for training, therefore,
a higher recognition accuracy is expected. But as observed, the performances
of disCNN flatten when more camera pairs are used, e.g. the same performance
under 18 and 56 camera pairs. One possible interpretation is: although
we have 56 camera pairs, the diagonal camera-pairs in the case of
56 pairs do provide new pose transformation information, since the
motion between a diagonal pair could be induced from motions of two
camera pairs in the case of 18 pairs, a horizontal camera pair and
a vertical camera pair.

%
 
\textbf{Qualitative visualizations}: Fig. \ref{fig:disentangling} shows k nearest neighbors of the query
image, based on the $\ell_{2}$ distances between their fc7-identity
(disCNN, 512D) and fc7 (AlexNet, 1024D) representations. We can see
clearly that disCNN successfully retrieves images of the same instance
under different poses as the nearest neighbors (Fig. \ref{fig:disentangling}
(a)). Although in some scenarios (Fig. \ref{fig:disentangling} (b)),
AlexNet find different images of the same instance as the nearest
neighbors, the retrieved neighbors clearly share similar poses as
the query image. These visualizations show disCNN disentangles
the representations of identity from pose, to some extent.

\noindent \textbf{Additional Notes}: we did do experiments on how the classification accuracy is affected by the size of the fc7 layer using AlexNet: given an AlexNet, we fix the architecture except the fc7 size, we set fc7 size to be {16,32,64,128,258,512,1024}, train them using 0.67 million images and evaluate the trained model on 0.22 million test images. The test accuracy increases as fc7 changes from 16 to 128, and then flattens from 128 all the way to 1024. This empirical results shows that: in our case to classify 10 categories, as long as the top layer reaches some size (128 in our case), then the network will have the capacity to categorize 10-class objects, by adding more neurons to fc7 does not improve the capacity of the network. This experiments indirectly shows: the superiority of disCNN over AlexNet is not because of the bottleneck representation of disCNN (1024D vs 512D in AlexNet).

\begin{table*}[!thbp]
\newcommand{\tabincell}[2]{\begin{tabular}{@{}#1@{}}#2\end{tabular}}
\centering{}%
\begin{tabular}{|c|c|c|c|c||c|c|c|c|c|c|}
\hline 
\# of camera pairs  & 3  & 6  & 9  & 12  & \# of camera pairs  & 3  & 6  & 9  & 12   \tabularnewline
\hline 
 \tabincell{c} {AlexNet \\(scratch)}  & 71.2  & 72.8  & 72.1  & 72.9  &  \tabincell{c}  {AlexNet \\ (AlexNet-iLab20M)}  & 76.2  & 77.3  & 79.9  & 79.6   \tabularnewline
\hline 
 \tabincell{c} {disCNN \\(scratch)}  & \textbf{75.0}  & \textbf{75.1}  & \textbf{77.0}  & \textbf{78.6}  &   \tabincell{c} {disCNN \\ (AlexNet-iLab20M)}  & \textbf{78.9}  & \textbf{80.8}  & \textbf{81.5}  & \textbf{82.7}   \tabularnewline
\hline 
\end{tabular}\caption{\label{tab:Object-recognition-rgbd} Object recognition accuracies ($\%$) of AlexNet and disCNN on the
Washington RGB-D dataset. The left (right) table shows performance comparisons between
disCNN and AlexNet trained from scratch (from the pretrained AlexNet features on the iLab-20M dataset).
As seen, by fine-tuning CNNs from features learned on iLab-20M, large performance gains are achieved, e.g. $\sim4.5\%$($\sim5.5\%$)
for disCNN (AlexNet). This shows features learned from iLab-20M are
effective for, and generalizable to object recognition in the RGB-D
dataset. Results from both tables shows disCNN outperforms AlexNet
by $\sim3.5\%$ (scratch) and $\sim2\%$ (fine-tune), which shows
the advantage of our disentangled architecture. Furthermore, when the number of camera pairs increases, the performances of
disCNN increase as well.}
\end{table*}

\begin{figure*}[!thbp]
\begin{centering}
\includegraphics[width=1\textwidth]{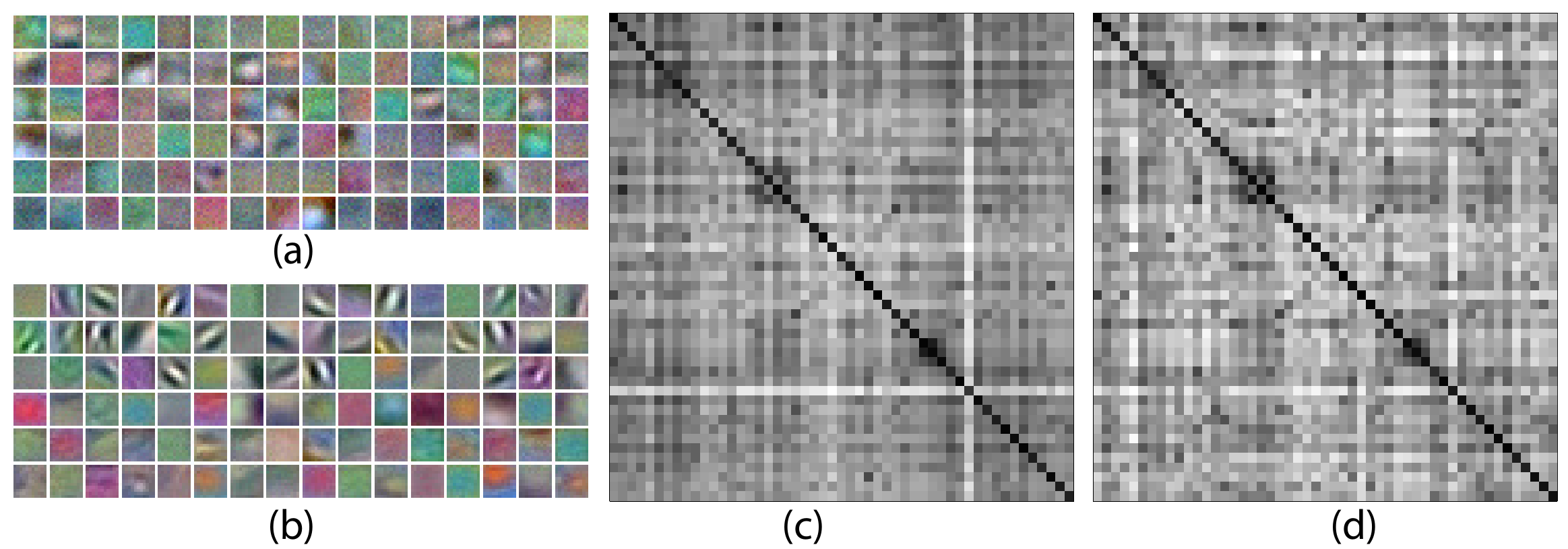} 
\par\end{centering}

\centering{}\caption{\label{fig:Learned-filters-and-l2-rgbd}Learned filters and between-class
$\ell_{2}$ distances. (a) and (b) show the learned filters of disCNN
trained from scratch, and disCNN fine-tuned from the pretrained AlexNet
on the iLab-20M dataset; (c) and (d) show the between-class $\ell_{2}$
distances of fc7 representations from AlexNet (1024D) and disCNN (512D).
Training disCNN from scratch learns only color blobs (a). (c,d) shows visually that
disCNN representations have smaller within category distances and
larger between category distances. The ratio between the mean between-category
distance and the mean within-category distance is 7.7/5.7 for disCNN/AlexNet.}
\end{figure*}

\subsection{Washington RGB-D dataset}

The RGB-D dataset \cite{lai2011large} depicts 300 common household objects organized into 51 categories. This dataset
was recorded using a Kinect style 3D camera that records synchronized and aligned 640x480 RGB and depth images at 30
Hz. Each object was placed on a turntable and video sequences were captured for one whole rotation.  For each object,
there are 3 video sequences, each recorded with the camera mounted at a different height so that the object is viewed
from different angles with the horizon. The dataset has a total of 250K images from different views and rotations. Two
adjacent frames have small motions, therefore are visually very similar, and in experiments, we pick one frame from each
5 consecutive frames, resulting in $\sim$50K image frames. Since the scale of the datasets does not match the scale of
ConvNets, we adopt the ``pretrain-finetuning'' paradigm to do object recognition in this dataset, using the pretrained
ConvNets weights on the iLab-20M dataset as initializations.

Training and test sets: \cite{lai2011large} provided 10 partition
lists of training/test. They use leave-one-out to partition:
randomly choose 1 instance within a category as test, and use the
remaining instances as training. Due to the training time limitation,
we evaluate performances using the first 3 partitions and report the
mean accuracies. We use the provided object masks to crop the objects
from the raw frames and resize them to the size 227$\times$227. Since objects are located
at the image center, by first cropping and then rescaling an image-pair
does not change the pose-transformation of the raw pair.

Camera pairs: similarly we take different numbers of camera-pairs and evaluate influence on the performances. In one
video sequence, every frame-pair with a fixed temporal gap could be imagined to be taken under a virtual camera-pair,
thus all such pairs share the same pose-transformation label. As an example, two pairs, $F_{i}-F_{i+\Delta}$ and
$F_{j}-F_{j+\Delta}$, whose temporal gap between frames are both $\Delta$, then they have the same pose-transformation
label. One $\Delta$ defines one camera-pair, and in experiments, we let $\Delta=\{5,10,15,20\}$.  Case one: we take
image-pairs with $\Delta=\{5\}$ from each video sequence, and all these pairs could be thought as taken by one virtual
camera pair, therefore have the same pose-transformation label. Since we have 3 video sequences, finally all pairs have
in total 3 pose-transformation labels, thus equivalently 3 virtual camera pairs; Case two: take image-pairs with
$\Delta=\{5,10\}$, end in 6 camera pairs; Case three: $\Delta=\{5,10,15\}$, end in 9 camera-pairs; case four:
$\Delta=\{5,10,15,20\}$, end in 12 camera-pairs. The total number of training image pairs under each case is 67K, 99K
and 131K respectively. The number of test images in all cases is 6773.

Implementation details: we use the same training settings as in iLab-20M
experiments to train AlexNet and disCNN, i.e., the same learning rates
(start from 0.01, end with 0.0001, with rate decreasing log linearly),
the same number of training epochs (15), and the same training order
within each epoch. We set $\lambda_{1}=1$ and $\lambda_{2}=0.05$
in experiments.

Results: we do two comparisons: first compare disCNN (AlexNet) trained from scratch against from the pretrained weights
on the iLab-20M dataset, then compare disCNN against AlexNet, both fine-tuned from the pretrained CNN features on
iLab-20M. Results are shown in Table \ref{tab:Object-recognition-rgbd}, our observations are: (1) disCNN (AlexNet)
trained by fine-tuning the pretrained AlexNet features on the iLab-20M wins over disCNN (AlexNet) trained from scratch
by $\sim4.5\%$($\sim5.5\%$), and their fine-tuned performances are better than the published accuracies, $74.7\%$, in
\cite{lai2011large} by a large margin. This shows the features learned from the iLab-20M dataset generalize well to the
RGB-D dataset; (2) disCNN outperforms AlexNet in both cases, either trained from scratch or from the pretrained AlexNet
features, which shows the superiority of the disentangling architecture over the linear chain, single task CNNs; (3)
similarly, we observe that the performance of disCNN increases as the number of camera pairs increase. We further
compute $\ell_{2}$ distances between categories using fc7-identity (disCNN, 512D) and fc7 (AlexNet, 1024D)
representations, and plot them in Fig. \ref{fig:Learned-filters-and-l2-rgbd}.  Visually the off diagonal elements in
disCNN are brighter and the diagonal elements are darker, showing smaller within-category distances and larger
between-category distances.

\subsection{ImageNet}

ImageNet has millions of labeled images, and training a ConvNet on a large dataset from pretrained models against from
scratch has been shown to have insignificant effects \cite{hinton2012deep,lecun2015deep}. In order to show that the
pretrained disCNN on the iLab-20M datasets learns useful features for object recognition, we fine-tune the learned
weights on ImageNet when only a small amount of labeled images are available. We fine-tune AlexNet using 5, 10, 20, 40
images per class (5K,10K,20K and 40K training images in total) from the ILSVRC-2010 challenge. AlexNet is fine-tuned
under three scenarios: (1) from scratch (random Gaussian initialization), (2) from pretrained AlexNet on iLab-20M, (3)
from pretrained disCNN on iLab-20M, and top-5 object recognition accuracies are presented in Table
\ref{tab:Top-5-object-recognition-ImageNet}. When we pretrain AlexNet and disCNN on the iLab-20M dataset, we use the
AlexNet with the number of neurons on the last two fully connected layers reset to 4096.

Results: (1) when only a limited number of labeled images are available,
fine-tuning AlexNet from the pretrained features on the iLab-20M dataset
performs much better than training AlexNet from scratch, e.g., the
relative improvement is as large as $\sim460\%$ when we have only
5 samples per class, and the improvement decreases when more labeled
images are available, but we still gain $\sim25\%$ improvements when
40 labeled images per class are available. This clearly shows features
learned on the iLab-20M dataset generalize to ImageNet. (2) fine-tuning
from the pretrained disCNN on iLab-20M performs even better than from
the pretrained AlexNet on iLab-20M, and this shows that disCNN learns
even more effective features for general object recognition than AlexNet.
These empirical results show the advantage of our disentangling architecture
to the traditional single task linear architecture.


\begin{table}[!htbp]
\newcommand{\tabincell}[2]{\begin{tabular}{@{}#1@{}}#2\end{tabular}}
\centering{}%
\begin{tabular}{|c|c|c|c|c|}
\hline 
\# of images/class  & 5  & 10  & 20  & 40 \tabularnewline
\hline 
\tabincell{c} {AlexNet \\ (scratch)}  & 1.47  & 4.15  & 16.45  & 25.89 \tabularnewline
\hline 
\tabincell{c} {AlexNet \\ (AlexNet-iLab20M)}  & 7.74  & 12.54  & 19.42  & 28.75 \tabularnewline
\hline 
\tabincell{c} {AlexNet \\ (disCNN-iLab20M)}  & \textbf{8.21}  & \textbf{14.19}  & \textbf{22.04}  & \textbf{30.19} \tabularnewline
\hline 
\end{tabular}\caption{\label{tab:Top-5-object-recognition-ImageNet}Top-5 object recognition
accuracies ($\%$) on the test set of ILSVRC-2010, with 150 images
per class and a total of 150K test images. First, fine-tuning AlexNet
from the pretrained features on the iLab-20M dataset clearly outperforms
training AlexNet from scratch, which shows features learned on the
iLab-20M dataset generalizes to ImageNet as well. Second, fine-tuning
from the pretrained disCNN-iLab20M performs even better than from
the pretrained AlexNet-iLab20M, which shows our disentangling architecture
learns even better features for object recognition than AlexNet.}
\end{table}


\section{Conclusions\label{sec:Conclusions}}

In this paper, we design a multi-task learning ConvNet to learn to
predict object categories. Unlike traditional ConvNets for object
recognition, which is usually a single task architecture and learns
features sensitive to the current task (i.e., object category) but
invariant to other factors of variation as much as possible (e.g.,
pose), disCNN retains all image generating factors of variation (object
category and pose transformation in our case), and learn them simultaneously
by explicitly disentangling representations of different factors.
Experiments on the large scale iLab-20M dataset show that features
learned by disCNN outperforms features learned by AlexNet significantly
for object recognition. If we fine tune object recognition on the
ImageNet dataset using pretrained disCNN and AlexNet features, disCNN-pretrained
features are consistently better than AlexNet-pretrained features.
All experiments show the effectiveness of our disentangled training
architecture.

As shown in \cite{agrawal2015learning}, features learned using egomotion
as supervision are useful for other vision tasks, including object
recognition, and the egomotion-pretrained features compare favorably
with features learned using class-label as supervision. In our paper,
we further showed that when our model has access to both object categories
and camera motions, it learns even better features than using only
class-label as supervision. One possible explanation is: although
egomotion learns useful features for object recognition, it does not
necessarily guarantee that feature representations of different instances
of the same class are similar since egomotion does not has access
to any class-label information. In our work, we showed, by feeding
ConvNets with additional class labels, the feature learning process
are further guided toward the direction that objects of the same class
tend to have spatially similar representations.

\noindent \textbf{Acknowledgement}: This work was supported by the National
Science Foundation (grant numbers CCF-1317433
and CNS-1545089), the Army Research Office (W911NF-
12-1-0433), and the Office of Naval Research (N00014-13-
1-0563). The authors affirm that the views expressed herein
are solely their own, and do not represent the views of the
United States government or any agency thereof.

{\small \bibliographystyle{ieee}
\bibliography{objectRecognition-twoStreams}

\begin{thebibliography}{10}\itemsep=-1pt

\bibitem{agrawal2015learning}
P.~Agrawal, J.~Carreira, and J.~Malik.
\newblock Learning to see by moving.
\newblock In {\em Proceedings of the IEEE International Conference on Computer
  Vision}, pages 37--45, 2015.

\bibitem{bengio2009learning}
Y.~Bengio.
\newblock Learning deep architectures for ai.
\newblock {\em Foundations and trends{\textregistered} in Machine Learning},
  2(1):1--127, 2009.

\bibitem{Borji_2016_CVPR}
A.~Borji, S.~Izadi, and L.~Itti.
\newblock ilab-20m: A large-scale controlled object dataset to investigate deep
  learning.
\newblock In {\em The IEEE Conference on Computer Vision and Pattern
  Recognition (CVPR)}, June 2016.

\bibitem{chen2014semantic}
L.-C. Chen, G.~Papandreou, I.~Kokkinos, K.~Murphy, and A.~L. Yuille.
\newblock Semantic image segmentation with deep convolutional nets and fully
  connected crfs.
\newblock {\em arXiv preprint arXiv:1412.7062}, 2014.

\bibitem{deng2009imagenet}
J.~Deng, W.~Dong, R.~Socher, L.-J. Li, K.~Li, and L.~Fei-Fei.
\newblock Imagenet: A large-scale hierarchical image database.
\newblock In {\em Computer Vision and Pattern Recognition, 2009. CVPR 2009.
  IEEE Conference on}, pages 248--255. IEEE, 2009.

\bibitem{dosovitskiy2015learning}
A.~Dosovitskiy, J.~Tobias~Springenberg, and T.~Brox.
\newblock Learning to generate chairs with convolutional neural networks.
\newblock In {\em CVPR}, pages 1538--1546, 2015.

\bibitem{girshick2014rich}
R.~Girshick, J.~Donahue, T.~Darrell, and J.~Malik.
\newblock Rich feature hierarchies for accurate object detection and semantic
  segmentation.
\newblock In {\em CVPR}, pages 580--587, 2014.

\bibitem{gkioxari2015contextual}
G.~Gkioxari, R.~Girshick, and J.~Malik.
\newblock Contextual action recognition with r* cnn.
\newblock In {\em Proceedings of the IEEE International Conference on Computer
  Vision}, pages 1080--1088, 2015.

\bibitem{hartley2003multiple}
R.~Hartley and A.~Zisserman.
\newblock {\em Multiple view geometry in computer vision}.
\newblock Cambridge university press, 2003.

\bibitem{he2015deep}
K.~He, X.~Zhang, S.~Ren, and J.~Sun.
\newblock Deep residual learning for image recognition.
\newblock {\em arXiv preprint arXiv:1512.03385}, 2015.

\bibitem{hinton2012deep}
G.~Hinton, L.~Deng, D.~Yu, G.~E. Dahl, A.-r. Mohamed, N.~Jaitly, A.~Senior,
  V.~Vanhoucke, P.~Nguyen, T.~N. Sainath, et~al.
\newblock Deep neural networks for acoustic modeling in speech recognition: The
  shared views of four research groups.
\newblock {\em Signal Processing Magazine, IEEE}, 29(6):82--97, 2012.

\bibitem{hinton2011transforming}
G.~E. Hinton, A.~Krizhevsky, and S.~D. Wang.
\newblock Transforming auto-encoders.
\newblock In {\em Artificial Neural Networks and Machine Learning--ICANN 2011},
  pages 44--51. Springer, 2011.

\bibitem{huang2013multi}
Y.~Huang, W.~Wang, L.~Wang, and T.~Tan.
\newblock Multi-task deep neural network for multi-label learning.
\newblock In {\em Image Processing (ICIP), 2013 20th IEEE International
  Conference on}, pages 2897--2900. IEEE, 2013.

\bibitem{ioffe2015batch}
S.~Ioffe and C.~Szegedy.
\newblock Batch normalization: Accelerating deep network training by reducing
  internal covariate shift.
\newblock {\em arXiv preprint arXiv:1502.03167}, 2015.

\bibitem{krizhevsky2012imagenet}
A.~Krizhevsky, I.~Sutskever, and G.~E. Hinton.
\newblock Imagenet classification with deep convolutional neural networks.
\newblock In {\em Advances in neural information processing systems}, pages
  1097--1105, 2012.

\bibitem{kulkarni2015deep}
T.~D. Kulkarni, W.~F. Whitney, P.~Kohli, and J.~Tenenbaum.
\newblock Deep convolutional inverse graphics network.
\newblock In {\em Advances in Neural Information Processing Systems}, pages
  2530--2538, 2015.

\bibitem{lai2011large}
K.~Lai, L.~Bo, X.~Ren, and D.~Fox.
\newblock A large-scale hierarchical multi-view rgb-d object dataset.
\newblock In {\em Robotics and Automation (ICRA), 2011 IEEE International
  Conference on}, pages 1817--1824. IEEE, 2011.

\bibitem{lecun2015deep}
Y.~LeCun, Y.~Bengio, and G.~Hinton.
\newblock Deep learning.
\newblock {\em Nature}, 521(7553):436--444, 2015.

\bibitem{lecun1998gradient}
Y.~LeCun, L.~Bottou, Y.~Bengio, and P.~Haffner.
\newblock Gradient-based learning applied to document recognition.
\newblock {\em Proceedings of the IEEE}, 86(11):2278--2324, 1998.

\bibitem{long2015fully}
J.~Long, E.~Shelhamer, and T.~Darrell.
\newblock Fully convolutional networks for semantic segmentation.
\newblock In {\em Proceedings of the IEEE Conference on Computer Vision and
  Pattern Recognition}, pages 3431--3440, 2015.

\bibitem{reed2014learning}
S.~Reed, K.~Sohn, Y.~Zhang, and H.~Lee.
\newblock Learning to disentangle factors of variation with manifold
  interaction.
\newblock In {\em Proceedings of the 31st International Conference on Machine
  Learning (ICML-14)}, pages 1431--1439, 2014.

\bibitem{seltzer2013multi}
M.~L. Seltzer and J.~Droppo.
\newblock Multi-task learning in deep neural networks for improved phoneme
  recognition.
\newblock In {\em Acoustics, Speech and Signal Processing (ICASSP), 2013 IEEE
  International Conference on}, pages 6965--6969. IEEE, 2013.

\bibitem{sermanet2013overfeat}
P.~Sermanet, D.~Eigen, X.~Zhang, M.~Mathieu, R.~Fergus, and Y.~LeCun.
\newblock Overfeat: Integrated recognition, localization and detection using
  convolutional networks.
\newblock {\em arXiv preprint arXiv:1312.6229}, 2013.

\bibitem{simonyan2014two}
K.~Simonyan and A.~Zisserman.
\newblock Two-stream convolutional networks for action recognition in videos.
\newblock In {\em Advances in Neural Information Processing Systems}, pages
  568--576, 2014.

\bibitem{simonyan2014very}
K.~Simonyan and A.~Zisserman.
\newblock Very deep convolutional networks for large-scale image recognition.
\newblock {\em arXiv preprint arXiv:1409.1556}, 2014.

\bibitem{soatto2016visual}
S.~Soatto, A.~Chiuso, and P.~Chaudhari.
\newblock Visual representations: Defining properties and deep approximations.
\newblock In {\em International Conference on Learning Representations},
  volume~3, 2016.

\bibitem{su2015render}
H.~Su, C.~R. Qi, Y.~Li, and L.~J. Guibas.
\newblock Render for cnn: Viewpoint estimation in images using cnns trained
  with rendered 3d model views.
\newblock In {\em Proceedings of the IEEE International Conference on Computer
  Vision}, pages 2686--2694, 2015.

\bibitem{szegedy2015going}
C.~Szegedy, W.~Liu, Y.~Jia, P.~Sermanet, S.~Reed, D.~Anguelov, D.~Erhan,
  V.~Vanhoucke, and A.~Rabinovich.
\newblock Going deeper with convolutions.
\newblock In {\em Proceedings of the IEEE Conference on Computer Vision and
  Pattern Recognition}, pages 1--9, 2015.

\bibitem{vedaldi2015matconvnet}
A.~Vedaldi and K.~Lenc.
\newblock Matconvnet: Convolutional neural networks for matlab.
\newblock In {\em Proceedings of the 23rd Annual ACM Conference on Multimedia
  Conference}, pages 689--692. ACM, 2015.

\bibitem{yang2015weakly}
J.~Yang, S.~E. Reed, M.-H. Yang, and H.~Lee.
\newblock Weakly-supervised disentangling with recurrent transformations for 3d
  view synthesis.
\newblock In {\em Advances in Neural Information Processing Systems}, pages
  1099--1107, 2015.

\bibitem{zhang2014improving}
C.~Zhang and Z.~Zhang.
\newblock Improving multiview face detection with multi-task deep convolutional
  neural networks.
\newblock In {\em Applications of Computer Vision (WACV), 2014 IEEE Winter
  Conference on}, pages 1036--1041. IEEE, 2014.

\bibitem{zhang2014facial}
Z.~Zhang, P.~Luo, C.~C. Loy, and X.~Tang.
\newblock Facial landmark detection by deep multi-task learning.
\newblock In {\em Computer Vision--ECCV 2014}, pages 94--108. Springer, 2014.

\bibitem{zhao2016improved}
J.~Zhao and L.~Itti.
\newblock Improved deep learning of object category using pose information.
\newblock In {\em arXiv preprint arXiv:1607.05836}, 2016.

\bibitem{zhu2014multi}
Z.~Zhu, P.~Luo, X.~Wang, and X.~Tang.
\newblock Multi-view perceptron: a deep model for learning face identity and
  view representations.
\newblock In {\em Advances in Neural Information Processing Systems}, pages
  217--225, 2014.

\end{thebibliography}
 } 
\end{document}